\documentclass{article}

\usepackage[preprint]{nips_2018}

\usepackage[utf8]{inputenc} % allow utf-8 input
\usepackage[T1]{fontenc}    % use 8-bit T1 fonts
\usepackage{hyperref}       % hyperlinks
\usepackage{url}            % simple URL typesetting
\usepackage{booktabs}       % professional-quality tables
\usepackage{amsfonts}       % blackboard math symbols
\usepackage{nicefrac}       % compact symbols for 1/2, etc.
\usepackage{microtype}      % microtypography
\usepackage{graphicx}
\usepackage{amsmath}
\usepackage{subcaption}
\usepackage{caption}

\title{
	At Human Speed: \\
	Deep Reinforcement Learning with Action Delay
}

\author{
    Vlad Firoiu\\
    DeepMind, MIT\\
    vladfi@google.com
  \And
    Tina W. Ju\\
    Stanford\\
    tinawju@stanford.edu
    \And
    Joshua B. Tenenbaum\\
    MIT\\
    jbt@mit.edu
}

\DeclareMathOperator*{\gru}{GRU}

\begin{document}

\maketitle

\begin{abstract}
There has been a recent explosion in the capabilities of game-playing artificial intelligence. Many classes of tasks, from video games to motor control to board games, are now solvable by fairly generic algorithms, based on deep learning and reinforcement learning, that learn to play from experience with minimal prior knowledge. However, these machines often do not win through intelligence alone -- they possess vastly superior speed and precision, allowing them to act in ways a human never could. To level the playing field, we restrict the machine's reaction time to a human level, and find that standard deep reinforcement learning methods quickly drop in performance. We propose a solution to the action delay problem inspired by human perception -- to endow agents with a neural predictive model of the environment which ``undoes’’ the delay inherent in their environment -- and demonstrate its efficacy against professional players in Super Smash Bros. Melee, a popular console fighting game.
\end{abstract}

\section{Introduction}

It has become ubiquitous to apply deep reinforcement learning methods to the games that humans enjoy. Perfect information games such as Go have fallen to a combination of deep RL and Monte-Carlo Tree Search \citep{silver2017mastering}, and even imperfect information games such as Poker are being solved \citep{deepstack}. Video games, starting with classic Atari console titles, were among the first to be tackled by deep RL (cite DQN), and are still widely used as benchmarks for state-of-the-art RL algorithms today. More recently, much interest has been shown in modern games such as StarCraft II \citep{starcraft2} and Dota 2 \citep{dota2}, which have established fan followings and professional scenes.

In all of these cases, the bar we wish our agents to reach is the level of competent or even world-class humans. This is especially true of those multi-player games in which humans can face off directly against trained AI opponents. It is certainly impressive and perhaps awe-inspiring to watch machines surpass us at the games that we have put in so much passion and dedication to master.

However, AI agents are often winning on more than intelligence alone -- they possess superhuman speed and precision by default. A more principled way to compare the intelligence -- that is, information processing -- abilities of machines and people would be to level the playing field in this regard. The addition of human constraints may also result in agents employing more interesting and relatable strategies to humans. 

To mimic the limits of human reaction time, we add fixed delay between the time an agent chooses an action and when that action reaches the environment.
To our knowledge, deep reinforcement learning methods have not been deliberately applied to 
environments with action delay. We investigate how deep RL methods perform with delay, and find that performance drastically falls as delay increases for agents playing Super Smash Bros. Melee and a variety of Atari 2600 games.

We present a novel technique for deep RL agents to cope with action delay, inspired by human perception and previous work on constant-delay Markov Decision Processes (MDPs). We endow agents with a neural predictive model of the environment, which can “undo” action delay, enabling them to act according to an estimate of the true state in which their action will be executed. Combining this predictive model with the IMPALA architecture, we extend the work in \citep{phillip} which trained superhuman SSBM undelayed agents via self-play. With this predictive architecture, agents are able to challenge world-class SSBM players while constrained by human-like reaction time.

% Existing RL methods <cite old paper> perform well but can't handle human-like delays due to <hypothesis>. We introduce a novel technique which achieves superhuman performance even under human-like constraints.

\section{Background}

\subsection{Super Smash Bros. Melee}

% here, talk about specifically what you're taking from phillip and why phillip is a great environment for trying this predictive modelling 
Super Smash Bros. Melee (SSBM) is a fast-paced multi-player fighting game released in 2001 for the Nintendo Gamecube. SSBM has steadily grown in popularity over its 17-year history, and today sports an active professional scene with tournaments that can draw hundreds of thousands of viewers. Although 2v2 matches are also played professionally, we focus on 1v1, which is the main tournament format.

We use the same interface to SSBM as in \citep{phillip}, which uses a discrete action set and structured state space with both discrete and continuous components.
While deep RL has often been applied to environments with visual state spaces such Atari \citep{ALE} and Deepmind-lab \citep{dmlab}, more recent work on Dota 2 and StarCraft II has used structured feature representations.
Rewards are given both for knock outs -- the underlying objective -- and damage, which is displayed on screen.

Being a fighting game, SSBM is naturally faster-paced than Dota or SC2. With important interactions occurring at such high frequency, human players are pushed to the limits of their reaction time. Without this handicap, relatively standard deep RL methods combined with self-play have surpassed human professionals \citep{phillip}. There even exists a hand-engineered decision tree-based AI which can play almost perfectly against humans, albeit in a limited setting where it can fully utilize unlimited reactions \citep{smashbot}. Given the importance of reaction time, SSBM is a natural environment in which to pose the problem of AI with action delay, from the point of view of both scientists and players.

\subsection{Delayed MDPs}

\citep{walsh2008} studied constant-delay Markov Decision Processes (CDMDPs), defined as MDPs where actions are delayed by a constant number of steps. They showed that state augmentation, which naively turns the CDMDP back into an MDP by appending the delayed actions to the state, is intractable due to the exponential blowup in the size of the new state space. They proposed Model-Based Simulation (MBS) as a sample efficient solution, similar to our approach, which is theoretically tractable when the underlying MDP is only “mildly stochastic”. Empirically, they found that MBS performs well on grid worlds, mazes, and the one-dimensional mountain car problem. We note that these environments are both simpler than SSBM and, crucially, are single-agent; the presence of an adversary greatly complicates the problem of modeling the environment.

\subsection{Reaction Time}

% TODO: cite these
% 247 \pm 18: https://www.ncbi.nlm.nih.gov/pmc/articles/PMC4456887/
% Dye, M. G. W., Green, C. S., & Bavelier, D. (2009). Increasing speed of processing with action video games. Current Directions in Psychological Science, 18, 321–326.

Fast-paced games like SSBM push players to the limits of their reaction time, which for the average person is about 250ms for visual stimuli \citep{reaction1}. It has been found that this reaction time both varies throughout the population and can be improved with training, such as by playing video games \citep{reaction2}. Human auditory reaction times are known to be somewhat faster, and indeed professional SSBM players will in certain situations listen for auditory cues instead of visual ones.

Many video games, Atari and SSBM included, run at 60Hz, which means that each frame lasts about 17ms. A completely undelayed agent thus has a reaction time of 17ms, while an agent under 15 frames of delay will have the reactions of an average human. We consider 12 frames to be the lowest human-plausible reaction time.

\section{Deep RL and action delay}

To our knowledge, deep reinforcement learning methods have not been deliberately applied to environments with action delay. \footnote{Anecdotally, we have heard that A3C performs significantly worse in OpenAI's Universe framework, which introduces a modest (40ms) length of delay.} That being so, an empirical investigation is in order.

\subsection{Setup}

For all experiments, we augment the environment with a length $d$ queue of actions. When the agent takes an action, it is pushed to the queue, and the action which pops out of the other end is executed instead. Thus, each action is executed exactly $d$ steps later than usual. Note that each step encompasses multiple game frames due to frame skipping.

The action queue is passed to the agent along with the state at each step, giving the agent in principle perfect information. This is known as the \emph{augmented} approach in \citep{walsh2008}.

\subsection{Atari}

\begin{figure}[ht]
\centering
\includegraphics[width=12cm]{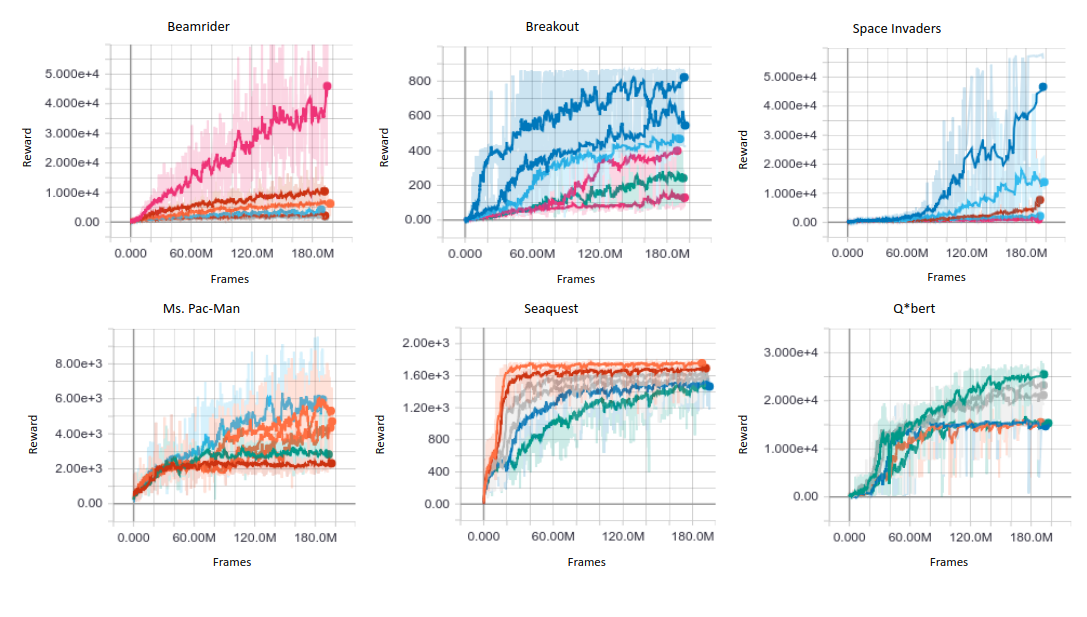}
\caption{IMPALA trained on Atari levels with delay varying between 0 and 5 (between 0ms and 333ms). For all games, final score was inversely correlated with delay.}
\label{fig:impala_atari}
\end{figure}

We trained IMPALA agents on six Atari games for 200 million frames using a frame skip of 4 and delays of 0 through 5 agent steps. Figure~\ref{fig:impala_atari} shows the learning curves of the agents with varied delay for each game. While the outcomes of Ms. Pacman were slightly mixed, increasing delay resulted in significantly lower scores on all other games.

\subsection{SSBM}

\begin{figure}[ht]
\centering
\includegraphics[width=12cm]{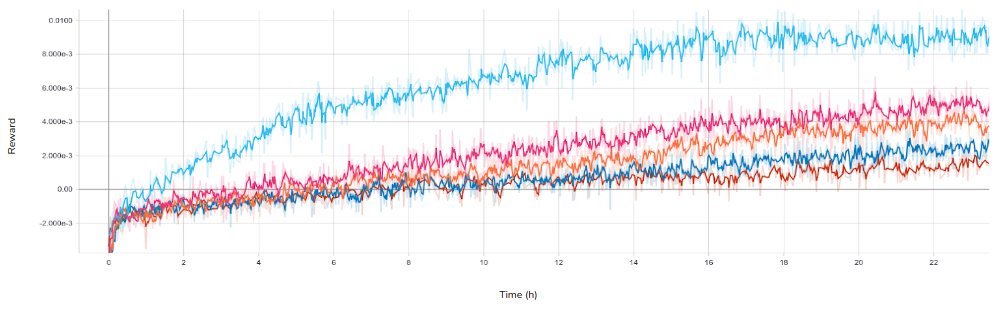}
\caption{Training against the in-game AI in SSBM with delays of 0 (light blue), 1 (magenta), 2 (orange), 4 (dark blue), and 5 (red) agent steps. Each step of delay measures 50ms. Learning speed and final rewards decrease significantly with increased delay.}
\label{fig:impala_smash}
\end{figure}

We trained IMPALA agents against the in-game AI at its hardest difficulty setting for one day using a frame skip of 3. Figure~\ref{fig:impala_smash} shows the learning curves of agents with varying delay against the in-game AI. Again, increasing delay dramatically lowered performance.

\subsection{Why is delay hard?}

\begin{figure}[ht]
    \centering
    \begin{subfigure}[b]{0.35\textwidth}
        \includegraphics[width=\textwidth]{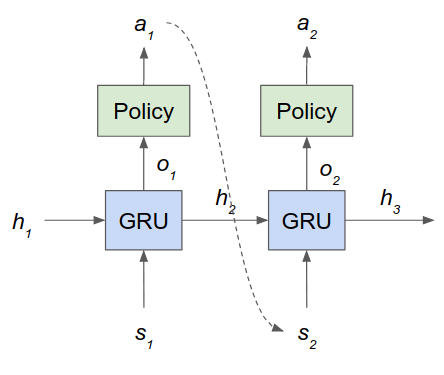}
        \caption{An agent unrolled over time.}
        \label{fig:traditional_nodelay}
    \end{subfigure}
    \begin{subfigure}[b]{0.583\textwidth}
        \includegraphics[width=\textwidth]{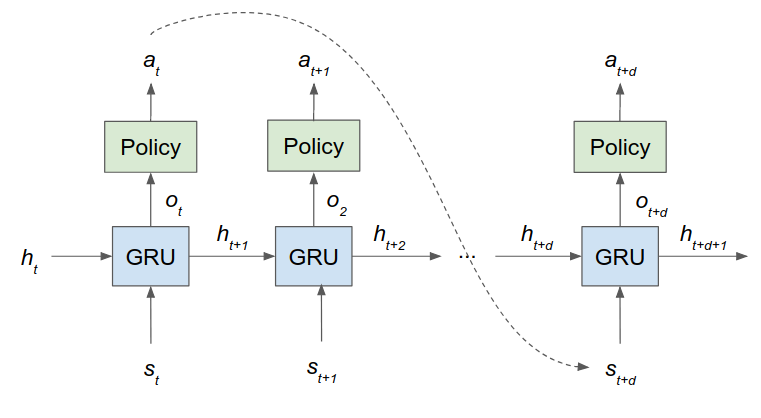}
        \caption{A delayed agent unrolled over time.}
        \label{fig:traditional_delay}
    \end{subfigure}
    \caption{Comparison of normal and delayed agent-environment interactions. }
    \label{fig:traditional_delay_archicture}
\end{figure}

As we have seen, agents under action delay perform quite poorly. Intuitively, we can see that, with delay, the agent does not know which state it will be in when its action is eventually executed by the environment, and without this knowledge it is difficult to act appropriately (Figure~\ref{fig:traditional_delay}), as compared to the process of an agent with no delay (Figure~\ref{fig:traditional_nodelay}).

This is especially problematic when it comes to the discrete components of the state, which can completely change the transition dynamics and therefore the optimal policy. For example, in SSBM each of the two characters has a discrete ``animation state'' which can take on over three hundred different values. Possible values discriminate between the twenty or so different attacks the character might be performing, whether the character is jumping, running, crouching, rolling, sliding, stunned from an enemy attack, and many others. Knowing which state your character is in is crucial for determining the best action.

Even the continuous components such as position can be tricky to deal with under uncertainty, as there is sharp discontinuity between an attack hitting or missing based on the distance of the characters.

More theoretically, we can measure the complexity of adding delay by considering the size of the resulting delayed MDP. In order to be Markovian, we must augment the original space $S$ with the queue of delayed actions $a_1, a_2, \ldots a_d \in A$. This results in an increase by a factor of $|A|^d$, which can easily become quite large.

\section{Predictive modeling as a solution to delayed actions}

\subsection{Human perception}

As we have seen, deep RL agents struggle in delayed environments. Since we wish to train policies that act under human-like delays, it is natural to ask how humans themselves deal with delay. Experimental psychology suggests that the brain constantly and subconsciously anticipates the near future in physical environments \citep{motion}. Optical illusions such as the Flash-Lag Effect show that our very perception of the present is actually a prediction, with moving objects placed in their extrapolated rather than present locations. This feature of our perceptual systems explains how we can perform athletic feats such as catching a baseball or returning a tennis serve with relatively slow motor controls.

\subsection{Predicting the present}
\begin{figure}[ht]
    \centering
    \begin{subfigure}[b]{0.7\textwidth}
        \includegraphics[width=\textwidth]{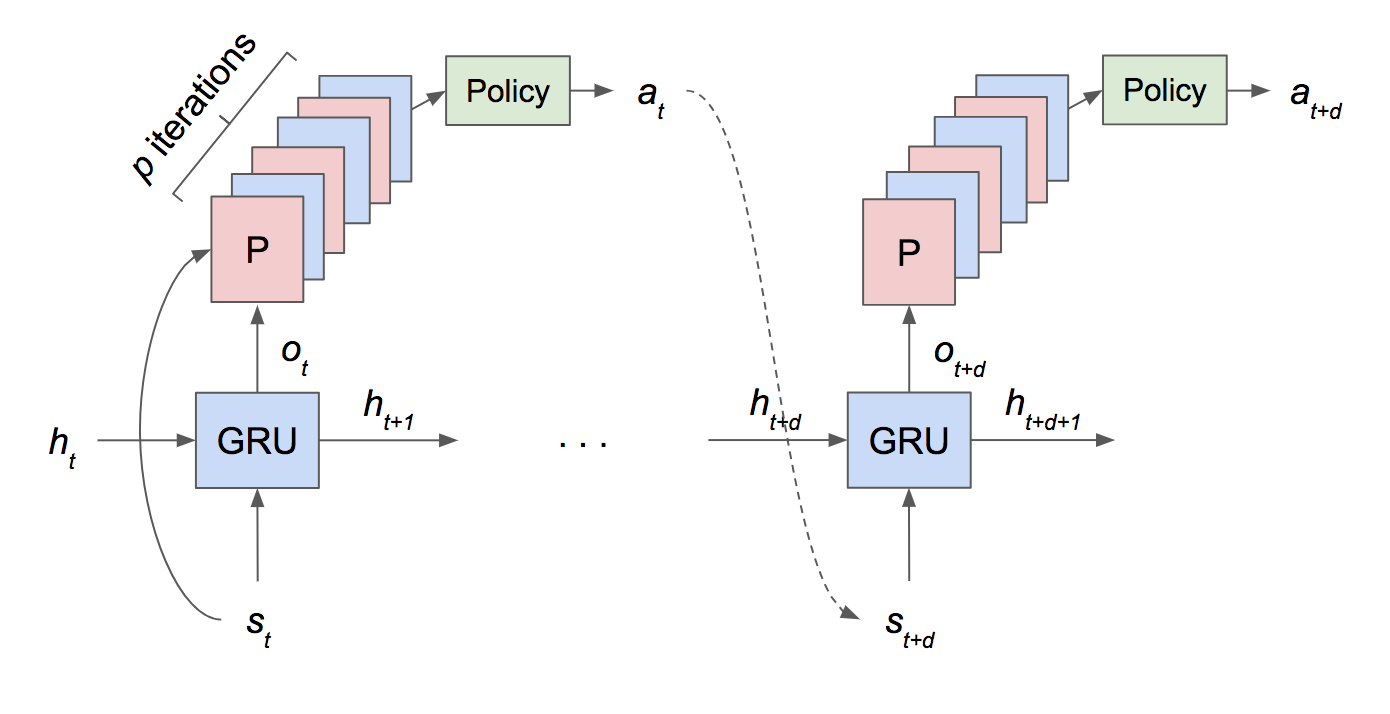}
        \caption{A deep RL agent with predictive model for coping with delay.}
            \label{fig:predictive}
    \end{subfigure}

	\begin{subfigure}[b]{0.7\textwidth}
        \includegraphics[width=\textwidth]{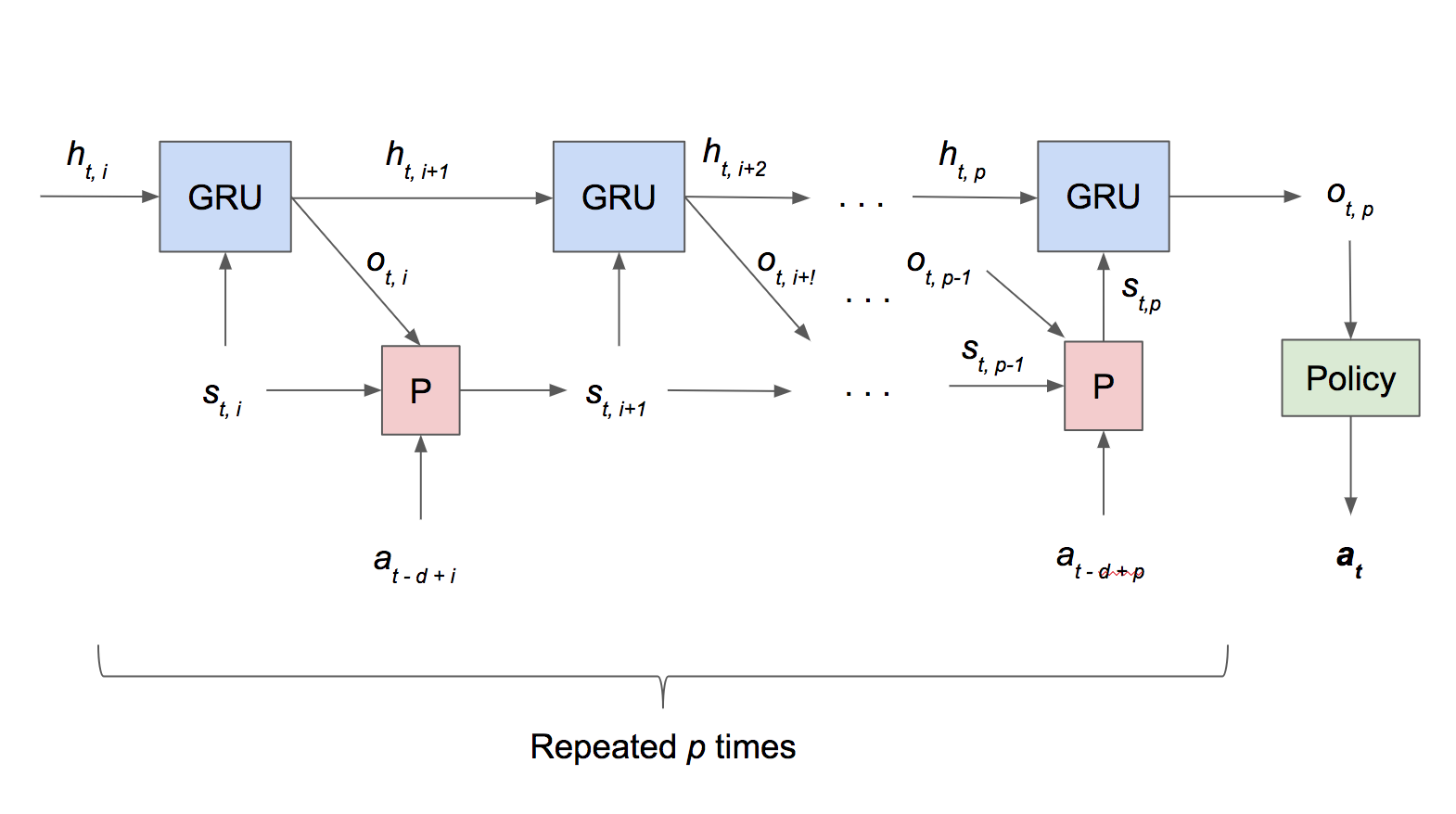}
        \caption{The predictive model unrolled over $p$ iterations to compute a single action.}
        \label{fig:predictive_unrolled}
    \end{subfigure}
    \caption{Illustration of the predictive architecture.}
    \label{fig:predictive_architecture}
\end{figure}

Taking this insight to heart, we endow our agents with a predictive model of the SSBM environment. Once trained, this model can be used to ``undo'' the agent's delay, as in MBS \citep{walsh2008}. Figure~\ref{fig:predictive_architecture} displays the predictive architecture, where Figure~\ref{fig:predictive} illustrates the predictive agent unrolled and Figure~\ref{fig:predictive_unrolled} shows the predictive model unrolled.

More precisely, suppose that $P(s, a)$ is the learned action-conditional transition model, the agent is under $d$ frames of delay, the current state is $s_t$, and the previously chosen actions were $a_{t-d}, a_{t-d+1}, \cdots a_{t-1}$. Due to the delay, the next action to be sent to the environment is precisely $a_{t-d}$, and the current decision $a_t$ will only be sent after state $s_{t+d}$. 

Our initial agents used a policy network that directly output $a_t$ given the augmented state $s_t, a_{t-d}, a_{t-d+1}, \cdots a_{t-1}$. With our predictive model, we can generate predicted states $s_{t, i}$ where

\begin{eqnarray*}
s_{t, 0} & =& s_t \\
s_{t, i+1} &=& P(s_{t, i}, a_{t -d + i})
\end{eqnarray*}

We say that a $(d, p)$ agent is one whose actions are under $d$ frames of delay and which runs the predictive model $p$ steps. In state $s_d$, the agent's policy network receives as input the predicted state $s_{d, p}$ and actions $a_p, a_{p+1}, \cdots a_{d-1}$.

Note that $d$ and $p$ are measured in the frames the agent sees, not counting those skipped. Thus, a $(d, p)$ agent acting every $f$ frames has a reaction time of $df$ frames. The frame skip itself adds another $(f-1)/2$ frames on average. When specifying the frame skip, refer to such an agent as a $(d, p, f)$ agent.

\subsection{Predictive architecture}

Our predictive model $P$ employs a residual-style architecture. 

\[
P(s, a) = F(s, a) \times (s + D(s, a)) + (1 - F(s, a)) \times N(s, a)
\]

Here:
\begin{itemize}
\item $D$ is a ``delta'' network which additively adjusts the previous state.
\item $N$ is a ``new'' network which constructs a new state.
\item $F$ is a ``forget'' network whose outputs are weights in [0, 1] and which smoothly interpolates between the adjusted and new states.
\item All three networks are feed-forward with output shapes equal to the state itself.
\item Addition and multiplication are done component-wise.
\end{itemize}

This architecture leverages the fact that our states $s$ are already encoded by semantically meaningful features. The changes in continuous components such as character position and velocity are well captured by the delta network. For the discrete components, we first transform from probability to logit space where addition is more meaningful. Interpreting the continuous components of the predicted state as means of fixed-variance normal distributions, the predicted state becomes a diagonal (that is, with independent components) approximation to the true distribution over states.

Although we omit their dependence on previous states, in practice the networks sit on top of a shared recurrent core using a Gated Recurrent Unit \citep{gru}. Using $h$ for core hidden states and $o$ for core outputs:

\begin{eqnarray*}
s_{t, 0} &=& s_t\\
{h_{t, 0}} &=& h_t\\
{h_{t, i+1}}, {o_{t, i}} &=& \gru({s_{t, i}}, {h_{t, i}})\\
{s_{t, i+1}} &=& P({s_{t, i}}, a_{t-d+i}, {o_{t, i}})\\
\end{eqnarray*}

\subsection{Training with delay}

We train our predictive model by regressing each predicted state $s_{t, i}$ to its true counterpart $s_{t+i}$. The distance between states is computed component-wise, with $L^2$ for the continuous components (character position, velocity, etc.) and cross-entropy for the discrete components.

Returns are computed somewhat differently for delayed agents. Because the action $a_t$ taken in state $s_t$ isn't executed until state $s_{t+d}$, it does not make sense to use any of the rewards $r_t, r_{t+1}, \ldots r_{t+d-1}$ for reinforcing $a_t$. Instead, we the return $R_{t+d} = r_{t+d} + \gamma r_{t+d+1} + \gamma^2 r_{t+d+2}\cdots$ from time step $t+d$, the point when $a_t$ is executed.
% * <zhukeepa@gmail.com> 2018-05-18T17:25:59.261Z:
% 
% use ?
% 
% ^.

This choice of return raises the question of what to do with the critic. Already, our objective has changed: at time $t$, we wish to estimate the expected return at time $t+d$ rather than time $t$. Intuitively, one might use the same predicted state $s_{t, d}$ that the policy does. However, because the critic is only used when training, we have full knowledge of the true state $s_{t+d}$, and so we can use that instead to form a more accurate value estimate. 

The policy gradient is largely unchanged, although one must be careful to compute the predicted state $s_{t, p}$ in the same manner on both the actor and learner. We found V-trace -- the off-policy correction algorithm introduced in \citep{impala} -- to be important, as the $p$ steps of prediction make the policy even more sensitive to changes in the parameters.

% equation putting everything together for policy gradient?

\subsection{Experiments}

\begin{figure}[ht]
    \centering
	\begin{subfigure}[b]{0.45\textwidth}
        \includegraphics[width=\textwidth]{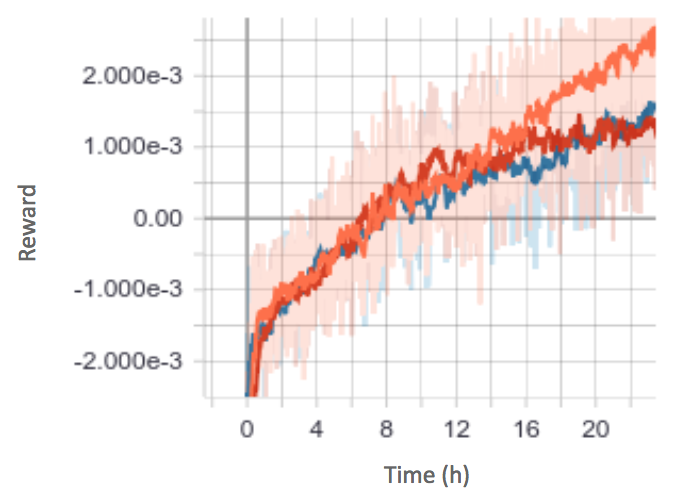}
        \caption{Predictive agents against the in-game AI: (4, 0) in orange, (4, 2) in blue, and (4, 4) in red.}
        \label{fig:predict_vs_cpu}
    \end{subfigure}
    ~ %add desired spacing between images, e. g. ~, \quad, \qquad, \hfill etc. 
    \begin{subfigure}[b]{0.45\textwidth}
		\includegraphics[width=\textwidth]{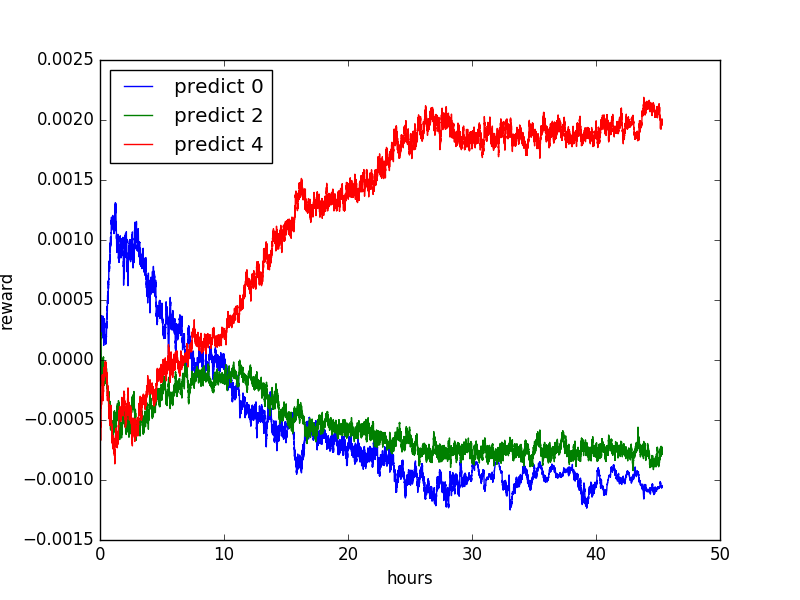}
		\caption{Average rewards for a population of agents playing against each other. The (4, 4) agent in red outperforms the (4, 2) in green and (4, 0) in blue by a wide margin.}
        \label{fig:predictive_vs_non}
    \end{subfigure}
\end{figure}

In the first test of our predictive architecture, we trained three agents: $(4, 0, 3)$, $(4, 2, 3)$, and $(4, 4, 3)$, against the in-game AI at its highest difficulty setting. As seen in Figure~\ref{fig:predict_vs_cpu}, we found the predictive agent to do slightly worse. Since the in-game AI is mostly deterministic and easily exploitable, and because the predictive model is non-trivially slower to run and train, against such a weak opponent the faster non-predictive agents can do slightly better in terms of wall-clock time.

Ultimately, performance against the in-game AI is not our real objective -- we wish to train agents with self-play that will be able to defeat human players. This suggests that we compare the predictive and non-predictive agents more directly, by having them train against each other. The resulting scores seen in Figure~\ref{fig:predictive_vs_non} clearly show the (4, 4) agent with a significant advantage over the other two, suggesting that the predictive model is necessary for learning more difficult policies. In particular, it appears that predicting only partially -- that is, with $p < d$ -- is insufficient, and best results are achieved with $p=d$.

Our final test was against ``Professor Pro'', the top player in the UK and ranked 41st internationally. To face him, we trained a (6, 6, 2) agent for three days, and then retrained it as a (7, 7, 2) agent for one week. Games were in tournament format -- first to four KOs -- and recorded at both delays 6 and 7. We also trained a non-predictive (6, 0, 2) agent for one week.

\begin{table}
  \caption{Performance of delayed agents against Professor Pro.}
  \label{pro-table}
  \vspace{.15cm}
  \centering
\begin{tabular}{ccccc} 
\toprule
\multicolumn{3}{c}{Agent} \\
\cmidrule{1-3}
Delay    & Prediction Steps & Days Trained & Wins & Losses \\
\midrule
6	&	0	&	7	&	0	& 6     \\
	&	6	&	3	&	3	& 5     \\
7	&	7	&	10	&	2	& 5     \\
\bottomrule
\end{tabular}
\end{table}

Although our predictive agents were not ultimately victorious, they did come close to even against a very skilled human opponent. We believe that with some additional work, perhaps by leveraging the predictive model for better exploration as in \citep{curiosity}, truly superhuman agents with human-level reactions will be possible.
% Say something here about the games!

\section{Future directions}

\subsection{Planning}

Perhaps the most promising extension of our work is to run the predictive model past the delayed action sequence and into the future. This opens the promising avenue of neural model-based planning that has proven immensely successful in perfect information games \citep{alphago}.

There are several challenges along this path, however. Without access to the true environment model, errors can quickly compound, making the resulting plan unreliable. This is exacerbated by the search procedure itself, which is likely to exploit flaws in the model as it tries to optimize reward. The approach taken in \citep{i2a} attempts to remedy this by allowing the policy to arbitrarily interpret the planned trajectory.

Another issue is runtime, which can be limited in real-time environments such as SSBM. Already, unrolling the predictive model can be quite expensive. While not an issue for a (7, 7, 2) agent, we found that at (9, 9, 2) the agent could not run quickly enough to keep up with a real-time environment, and thus could not play against human opponents. However, there are certainly opportunities for improving the model’s computational complexity, for example by precomputing predictive steps before they are needed.

\subsection{Modeling the opponent}

While we demonstrate that our approach can perform well in the multi-agent setting -- that is, when the opponent is also learning -- our predictive model ignores the opponent, effectively pretending that the opponent is a part of the environment. With privileged post-facto information of the opponent’s actions, one could train a model that conditions on both players’ actions, and use it to reason about the underlying imperfect-information game. In this form it would be possible to apply methods from \citep{deepstack}, though to our knowledge this has yet to be attempted with a neural environment model.

\subsection{Other temporal action spaces}

While constant delay may be a reasonable proxy for human reaction time, in other contexts such as robotics (especially over an unreliable network) variable delay may be more accurate. Constructing models that can deal with variable delay in real time is likely to be difficult, and it may be more pragmatic to simply move to lower-frequency policies.

Another limitation that humans have, aside from reaction time, is their total number of actions per minute (APM). Even in games such as StarCraft which are known for high APM, top professionals rarely exceed 400 APM, well below the 1800 taken by an RL agent with frame skip of two. Clearly humans are being much more efficient, acting only when it is truly necessary to do so. An RL agent that could decide not to act might even learn more effectively, as the credit assignment problem becomes easier when there are fewer actions that need to be reinforced.

\section{Conclusion}

In this paper we consider the problem of deep reinforcement learning in environments with action delay. We find that standard methods such as IMPALA are ill-equipped to deal with this new challenge and rapidly lose performance with increasing delay. Inspired by human visual perception and previous work on constant-delay MDPs, we propose a solution using a predictive environment model to anticipate the future state on which the current action will act. This provides the right inductive bias that is missing from the simpler augmented-state approach, endowing the agent with a model that more closely matches reality. Empirically, we find that predictive agents significantly outperform non-predictive ones when matched head to head, and can even hold their own against highly-ranked human professionals.

\bibliographystyle{named}
\bibliography{new-arxiv}

\end{document}